\RequirePackage{amsmath}

\documentclass[runningheads]{llncs}

\usepackage[T1]{fontenc}
\usepackage{graphicx}

\usepackage[utf8]{inputenc} % allow utf-8 input
\usepackage[T1]{fontenc}    % use 8-bit T1 fonts
\usepackage{hyperref}       % hyperlinks
\usepackage{url}            % simple URL typesetting
\usepackage{booktabs}       % professional-quality tables
\usepackage{amsfonts}       % blackboard math symbols
\usepackage{nicefrac}       % compact symbols for 1/2, etc.
\usepackage{microtype}      % microtypography
\usepackage{lipsum}
\usepackage{fancyhdr}       % header
\usepackage{graphicx}       % graphics
\graphicspath{{media/}}     % organize your images and other figures under media/ folder
\usepackage{enumerate}
\usepackage{mathtools}
\usepackage{float}
\usepackage{pifont}

\usepackage{amsopn, amssymb, amscd, amsmath}

\usepackage[ruled]{algorithm2e}

\usepackage{tikz-cd}

\usepackage{physics}

\usepackage{bbm}

\DeclareMathOperator{\IG}{IG}

\providecommand{\dd}{\mathrm{d}}

\providecommand{\R}{\mathbb{R}}

\begin{document}
\title{Graph-based Integrated Gradients for Explaining Graph Neural Networks}
\titlerunning{Graph-based Integrated Gradients}
% \author{Lachlan Simpson\inst{1} \and
% Kyle Millar\inst{2}
% \and
% Adriel Cheng\inst{1,2} \and Cheng-Chew Lim \inst{1} \and Hong Gunn Chew \inst{1}}
%
% \authorrunning{L. Simpson et al.}
% \institute{The University of Adelaide
% \\
% \and
% Defence Science and Technology Group}
\author{Lachlan Simpson\inst{1}\and
Kyle Millar\inst{2}
\and
Adriel Cheng\inst{1,2} \and Cheng-Chew Lim \inst{1} \and Hong Gunn Chew \inst{1}}

\authorrunning{L. Simpson et al.}
\institute{School of Electrical and Mechanical Engineering, The University of Adelaide
\\
\and
Defence Science and Technology Group}
\maketitle              % typeset the header of the contribution
\begin{abstract}

\end{abstract}

\begin{abstract}
Integrated Gradients (IG) is a common explainability technique to address the black-box problem of neural networks. Integrated gradients assumes continuous data. Graphs are discrete structures making IG ill-suited to graphs. In this work, we introduce graph-based integrated gradients (GB-IG); an extension of IG to graphs. We demonstrate on four synthetic datasets that GB-IG accurately identifies crucial structural components of the graph used in classification tasks. We further demonstrate on three prevalent real-world graph datasets that GB-IG outperforms IG in highlighting important features for node classification tasks. 
\end{abstract}

\section{Introduction}
Modelling how entities in data are related to one another is a primary focus in biology, computer science, and the social sciences \cite{Newman2010}. Protein interactions \cite{Jha_Saha_Singh_2022}, malicious activity on a computer network \cite{9737249} and recommending products \cite{FOULADVAND2023104407} are examples. Such relationships within data are aptly described by graphs \cite{Newman2010}.

The efficacy of graph neural networks in graph learning tasks comes with the trade-off of explainability \cite{zednik2019solving}. The limited explainability of neural networks results in a lack of trust in predictions \cite{Hallowell_Badger_Sauerbrei_Nellåker_Kerasidou_2022}. The increasing distrust in neural networks in safety critical systems, along with the EU AI safety act and Australian AI safety guardrails, necessitate the development of graph explainability models \cite{AUS2025AIGuardrails,EU2025AIAct}. Classical explainable AI (XAI) on continuous array-based data provides an explanation of a prediction in terms of the input features. Graph explainability is fundamentally different as an explanation requires both features and structure. Geometric deep learning (GDL) introduced by Bronstein et al. \cite{bronstein2021geometricdeeplearninggrids} demonstrates that incorporating the geometry (structure) of data is crucial in developing robust machine learning techniques. The application of GDL to XAI has proven effective in developing adversarial and user-friendly XAI techniques \cite{10935253,simpson-basepoint}.

Integrated gradients \cite{pmlr-v70-sundararajan17a} is a prevalent XAI technique deployed ubiquitously to explain the predictions of machine learning models. Recent work has focused on extending IG to datasets with continuous non-Euclidean geometry \cite{costanza2025,sanyal-ren-2021-discretized,simpson-basepoint,zaher2024manifoldintegratedgradientsriemannian}. There is yet to be an extension of IG to discrete data such as graphs. In this work using the principals of GDL, we propose graph-based integrated gradients(GB-IG); a novel extension of IG to graphs. We demonstrate that GB-IG yields explanations that successfully highlight important nodes and geometric structures in classification problems which classical IG fails to.

The contributions of this work are twofold:
\begin{enumerate}
    \item We introduce graph-based integrated gradients (GB-IG), a novel extension of IG which outperforms IG in identifying important nodes and substructures in node classification tasks.
    \item We demonstrate that GB-IG satisfies several theoretical axioms of IG. We introduce a new axiom, path-based completeness. Path-based completeness reduces to regular completeness in $\R^n$. The axioms ensure an implementation of GB-IG may be rigorously validated.
\end{enumerate}
The remainder of this work is structured as follows: Section 2 discuses related work, Section 3 provides a background on graph machine learning and integrated gradients, Section 4 introduces GB-IG, Section 5 compares GB-IG with variants of IG across four synthetic and three real-world graph datasets. We conclude in Section 6 with a discussion of future works. 

\section{Related Work}
Incorporating the geometry of the data into integrated gradients is done within the context of the manifold hypothesis. The manifold hypothesis stipulates that datasets lie on a low-dimensional Riemannian manifold \cite{fefferman2013testing,whiteley2024statisticalexplorationmanifoldhypothesis}. We note that work on incorporating geometry into XAI models utilise image classification datasets, not graphs. Extending IG to incorporate geometry is undertaken within a Riemannian framework. Zaher et al. \cite{zaher2024manifoldintegratedgradientsriemannian} propose manifold integrated gradients (MIG). The shortest path between two points in Euclidean space is a straight line. On a Riemannian manifold the shortest path is a length minimising geodesic. MIG replaces the straight line equation in IG with a geodesic on the manifold. The geodesic ensures the explanations remain within the distribution of the data. 

In \cite{simpson-basepoint} the authors leverage the tangent space of the Riemannian manifold to construct a base-point such that the explanations align with the tangent space. Tangentially aligned integrated gradients ensures the explanations are meaningful to the user and highlight the object to be classified \cite{brodt}. In the domain of word embeddings Sanyal et al. \cite{sanyal-ren-2021-discretized} propose discretising IG along non-linear paths in the embedding space. Discretised IG (DIG) corresponds to the discrete nature of word embeddings. Costanza et al. \cite{costanza2025} introduce Riemannian IG (RIG). RIG abstracts integrated gradients to a general compact connected Riemannian manifold, not necessarily embedded in Euclidean space.

The extension of IG to a smooth manifold is the central theme in the aforementioned techniques. Whilst DIG discretises IG to non-linear paths, the points are embedded in a continuous space, namely the embedding manifold. In this work, we extend both MIG and DIG to undirected and unweighted graphs. MIG demonstrates that in non-Euclidean geometry, the shortest paths between two points is not necessarily a straight line. For a graph equipped with a distance metric the shortest path is not a straight line as there is no notion of a straight line in a graph nor is the shortest path unique \cite{Newman2010}. DIG provides the methodology to construct IG along non-linear paths. Graphs may be considered as discretised manifolds \cite{Ni_Lin_Luo_Gao_2019}. GB-IG is then a discretised extension of MIG and DIG, a central contribution of this work.

\section{Graph Machine Learning and Integrated Gradients}

Graphs consist of a set of nodes $\mathcal{N}$ and edges $\mathcal{E}$. The set of edges $\mathcal{E}$ defines the connections between nodes and may be represented by the adjacency matrix $\mathbf{A}$. The adjacency matrix is an $|\mathcal{N}| \times |\mathcal{N}|$ binary matrix where $\mathbf{A}_{ij} = 1$ if $(i,j) \in \mathcal{E}$ and $0$ otherwise. Nodes may be equipped with $d$-dimensional feature vectors. The set of feature vectors is an $|\mathcal{N}| \times d$ matrix denoted by $\mathbf{X}$. We will denote a graph as the pair $\mathcal{G} = (\mathbf{X},\mathbf{A})$. Graph neural networks operate on the assumption that a node is characterised by its relationship to neighbouring nodes. Given a target node, the feature vectors of neighbouring nodes are aggregated together to form a representation of the target node. Given a graph $(\mathbf{X},\mathbf{A})$, the $l$-th graph convolutional network (GCN) layer first proposed by Kipf and Welling \cite{kipf2017semi} is defined as: 
\begin{equation}
\label{eqn:GCN}
    \mathbf{H}^{(l+1)} = \sigma(\Tilde{\mathbf{D}}^{-\frac{1}{2}}\Tilde{\mathbf{A}}\Tilde{\mathbf{D}}\mathbf{H}^{(l)}\mathbf{W}^{(l)}),
\end{equation}
where, $\mathbf{H}^{(0)} = \mathbf{X}$, $\Tilde{\mathbf{A}} = \mathbf{A}+\mathbf{I}$, $\Tilde{\mathbf{D}}$ is the diagonal degree matrix of $\Tilde{\mathbf{A}}$, $\mathbf{W}^{(l)}$ is a learnable weight matrix, and $\sigma$ is a non-linear activation function. Equation \ref{eqn:GCN} demonstrates that the structure and features of a graph are coupled together in graph neural networks. 
Integrated gradients (IG) is a popular gradient explainability technique. IG is defined below.

\begin{definition} \cite{pmlr-v70-sundararajan17a}
Given a base-point $b \in \R^d$ and a point to explain $x \in  \R^d$, we define IG as
\begin{equation}
    \IG(x,b) = (x-b) \odot \int_{0}^{1} \nabla f(tx+(1-t)b) \dd t,
\end{equation}
where $\odot$ is the Hadamard product.
\end{definition}
IG assumes data is continuous and the shortest path between two points is a straight line. The base-point is chosen a-priori and fundamentally alters the explanations \cite{kindermans2017unreliability}. Below we provide three commen base-point choices for IG in literature \cite{sturmfels2020visualizing}: 

\begin{enumerate}
    \item \textbf{Zero}. Here the base-point for all points is a constant zero vector
    \begin{equation}
        \label{eqn:zero}
        b^{\text{zero}} = 0.
    \end{equation}
    
    \item \textbf{Uniform}. We sample uniformly over a valid range of $\mathbf{X}$
    \begin{equation}
    \label{eqn:uniform}
    b^{\text{uniform}}_{i} \sim U(\min_{i},\max_{i}).
    \end{equation}

    \item \textbf{Gaussian}. A Gaussian filter is applied to the input $x$. 
    \begin{equation}
        \label{eqn:gauss}
        b^{\text{Gaussian}} = \sigma \cdot v+x,
    \end{equation}
    where, $v_{i} \sim \mathcal{N}(0,1)$ and $\sigma \in \mathbb{R}$. 
\end{enumerate}
For a discussion of the impact of each base-point choice we refer the reader to \cite{simpson-basepoint}.

\begin{figure}[!bp]
    \centering
    \includegraphics[width=0.6\linewidth]{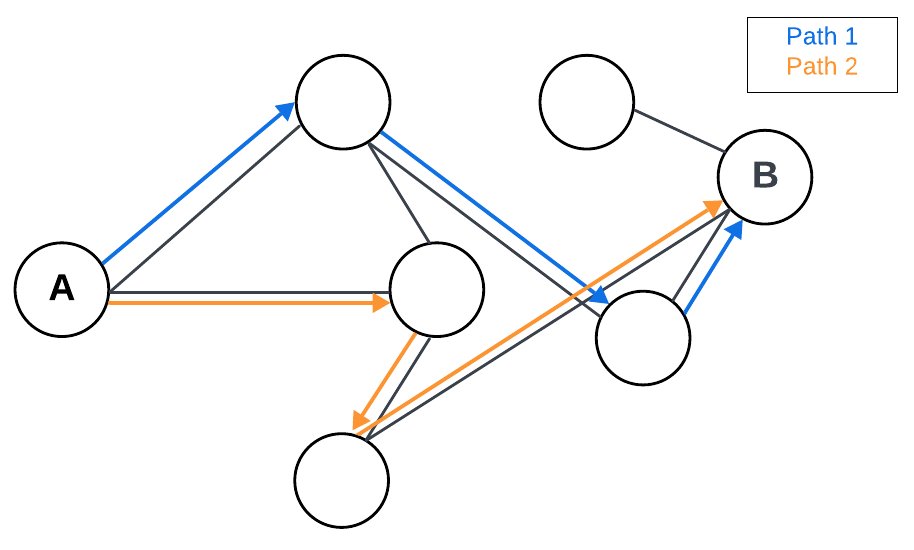}
    \caption{Example undirected graph with multiple shortest paths. Node A is the node to explain and node B is the base-point. Shortest paths between nodes A and B are indicated by blue and orange paths.}
    \label{fig:enter-label}
\end{figure}
\section{Graph-Based Integrated Gradients}
Paths on a graph must be defined to incorporate the structure of a graph into IG. Given an unweighted graph $\mathcal{G}$, a path between two nodes $x,y \in \mathcal{G}$ is a sequence of nodes connecting $x$ and $y$. The length of a path is the number of edges. Unlike a straight line in $\R^n$, the shortest path on a graph is not unique \cite{Newman2010}. Figure~\ref{fig:enter-label} provides an example graph where two shortest paths exist. IG accumulates the attribution over a path between a base-point and the point to explain. In Figure~\ref{fig:enter-label} the explanations are accumulated over both paths. The explanations of both paths are summed together to provide the mean contribution of structure.

Equation \ref{eqn:GCN} demonstrates that a GCN couples the adjacency and feature matrices together in creating node representations. We may think of the features as embedded within the graph space. Standard IG (Definition 1) interpolates the feature matrix $\mathbf{X}$ along a straight line from a base-point. 
A straight line interpolation is not defined when considering the geometry of a graph. In the following, we extend integrated gradients to graphs.

Let $\Gamma(b, x) = \{ \gamma \colon [0,1] \to G \mid \gamma(0) = b$, $\gamma(1) = x$\} consist of the shortest paths between nodes $b,x \in G$. Note that paths are finite sequences of nodes. Given a path $\gamma \in \Gamma(b, x)$, we will extend IG to graphs with the following equation
\begin{equation}
\label{eqn:dig}
    \int_{\gamma} \frac{\partial F(\mathbf{X},\mathbf{A})}{\partial x_{j}} \dd t \approx \sum^{|\gamma|}_{i} [\mathbf{X}(\gamma_{i+1})-\mathbf{X}(\gamma_{i})]_{j}\frac{\partial F(\mathbf{X}(\gamma_{i}),\mathbf{A})}{\partial x_{j}},
\end{equation}
where, $X(\gamma_{i})$ denotes the nodal-feature of the i-th node $\gamma_i$ on the path. Equation~\ref{eqn:dig} is DIG applied to graphs, where instead of considering the difference between word embeddings, we consider the difference between nodal attributes. Equation~\ref{eqn:dig} accounts for both impact of features on the prediction (the partial derivative) and the structure (difference in the nodal attributes along a path). The following equation accounts for the contribution of each shortest path
\begin{equation}
    \text{GB-IG}_{j}(b,x) = \sum_{\gamma \in \Gamma}\frac{1}{|\gamma|}\sum^{|\gamma|}_{i} [\mathbf{X}(\gamma_{i+1})-\mathbf{X}(\gamma_{i})]_{j}\frac{\partial F(\mathbf{X}(\gamma_{i}),\mathbf{A})}{\partial x_{j}}.
\end{equation}
Choosing a base-point $b$ is an open problem for IG \cite{drakard2022exploring,simpson-basepoint}. The zero base-point $b = 0$ is the common choice in machine learning \cite{NEURIPS2020_417fbbf2,pmlr-v196-xenopoulos22a}. Interpolating the feature matrix from $\mathbf{0} \to \mathbf{X}$ is ill-posed for a graph structure as a straight line is not defined in a discrete graph space.

Another common choice of base-point is the maximum distance, where, for a given point $x$ one takes $b = \mathrm{argmax}_{y} \|y-x\|_{p}$. We adapt the maximum distance base-point to graphs where we take the base-point to be the node with maximal distance from the node $x$. Maximal distance is not necessarily unique. Consider for example, the star graph of $n$ nodes $S_n$. Every node from the center is of maximal distance to the central node. Let $D(x)$ denote the set of nodes of maximal distance from $x$. GB-IG with the maximal distance base-point becomes:
\begin{equation}
    \sum_{b \in D(x) }\text{GB-IG}_{j}(b,x).
\end{equation}
Considering all points in $D(x)$ introduces a large computational complexity, in the following section we introduce an information theoretic approach to choose a single base-point in $D(x)$.
 
\subsection{Base-point Selection: An Information-Theoretic Approach}
The number of maximal distance nodes can in the worst case be $\mathcal{O}(|V|-1)$. One would then need to consider all shortest paths from the maximal distance nodes. Here, we propose an information-theoretic approach to select a single base-point reducing computational complexity.

GNNs are message-passing networks \cite{kipf2017semi}. The path which contains the highest ability to propagate information is structurally important. Given a path $\gamma \in \Gamma(x,y)$ we define the information of $\gamma$ as:
\begin{equation}
    I(\gamma) = -\log_2(\mathrm{p}(\gamma)) \coloneqq -\log_2\left(\prod_{i = 0}^{|\gamma|} \frac{1}{\deg(\gamma_i)}\right),
\end{equation}
where, $\deg$ denotes the degree of the node. The above reduces to
\begin{equation}
    I(\gamma) = \sum_{i = 0}^{|\gamma|}\log_2(\deg(\gamma_i)).
\end{equation}
Paths with high connectivity therefore carry the most information. The entropy $E(\Gamma)$, of a set of paths $\Gamma$ is defined as:
\begin{equation}
    E(\Gamma) = \sum_{\gamma \in \Gamma}\mathrm{p}(\gamma)I(\gamma).
\end{equation}
Given a set of maximal distance base-points $D(x) \subseteq \mathcal{N}$ we choose the optimal base-point by
\begin{equation}
    \Tilde{b} = \mathrm{argmax}_{b \in D(x)} E(\Gamma(b,x)).
\end{equation}
Whilst the above approach reduces the number of potential base-points from $|\mathcal{N}|-~1$ to a single point $\Tilde{b}$, the number of paths $|\Gamma(\Tilde{b},x)|$ may be quite large. Reducing the number of paths in $\Gamma(\Tilde{b},x)$ will improve computational and explanation complexity. Explanation complexity will be defined and discussed in Section 5.

Variants of IG seek to satisfy six desirable axioms proposed by Sundararajan et al. \cite{pmlr-v70-sundararajan17a}. We refer the reader to \cite{lundstrom2022rigorous} for an in-depth discussion of the axioms. Below we introduce each axiom and prove GB-IG satisfies axioms 1-5. Axiom 6 completeness is extended to graphs. We show GB-IG satisfies path-wise completeness and in the case of Euclidean space GB-IG satisfies regular completeness.

\subsection{Axioms Satisfied by GB-IG}

Given a DNN $F$, a point to explain $x$ and a base-point $b$ which all explanations are made relative to, Sundararajan et al. \cite{pmlr-v70-sundararajan17a} introduce six axioms of a gradient explainability model $A(x,b,F)$. The axioms, whilst having interesting theoretical applications, allow an implementation of an explainability technique to be sanity checked. The axioms are:

\begin{enumerate}
    \item Nullity.

    If $\frac{\partial F}{\partial x_i} = 0$ then
    \begin{equation}
        A_i(x,b,F) = 0.
    \end{equation}
    
    Features which makes no contribution to the model have an attribution of zero.
    
    \item Implementation Invariance. 
    
    If two models are functionally equivalent i.e. $F(x) = G(x) \; \forall \: x$, then
    \begin{equation}
    A(x,b,F) = A(x,b,G) \text{ for all } x,b \in \R^n.   
    \end{equation}
    
    \item Linearity.
    \begin{equation}
         A(x,b,\alpha F+ \beta G) = \alpha A(x,bF) + \beta A(x,b,G).   
    \end{equation}
    Linearity ensures aggregated models also have aggregated attributions. Furthermore, linearity with respect to scalars ensures that if a model is scaled so to is the attribution. 

    \item Sensitivity.

    Suppose $x_i = x_i'$, $x_j \neq x_j' \: \forall i \neq j$ and $F(x) \neq F(b)$. Then
    \begin{equation}
        A_i(x,b, F) \neq 0.
    \end{equation}
    \item Symmetry.

    Suppose $F(x)$ is invariant under permutation of the input (i.e. $F(x,y) = F(y,x)$). Then so to is $A(x,b,F)$.

     \item Completeness.
    
    The sum of the attributions is equal to the difference between the model evaluated at the input $x$ and the base-point $b$. 
    \begin{equation}
     \sum_{i}^{n}A_{i}(x,b,F) = F(x)-F(b).   
    \end{equation}
\end{enumerate}
Axioms 1-5 are immediately satisfied. GB-IG does not satisfy completeness as the attributions are accumulated over all paths. GB-IG is path-wise complete. 

\begin{proof}
\begin{align*}
      \sum_j  \text{GB-IG}_{j}(\Tilde{x},x)
      &= \sum_j \sum_{\gamma \in \Gamma} \int_{\gamma} \partial_{j}F \dd t\\
      &= \sum_{\gamma \in \Gamma} \sum_{j} \int _{\gamma} \partial_{j}F \dd t\\
      &= \sum_{\gamma \in \Gamma} F(\gamma(1))-F(\gamma(0)).
\end{align*}
When $\Gamma$ consists of one path, we recover standard completeness in Euclidean space.
\end{proof}
\section{Experiments}

In this section GB-IG is compared with IG under varying base-point choices on four synthetic and three standard graph ML datasets. Three standard metrics: fidelity \cite{yuan2023explainability}, sparsity \cite{Pope_2019_CVPR}, and the Jaccard index \cite{taha2015metrics} evaluate the performance of the graph XAI techniques. A three-layer GCN with 64 nodes in each hidden layer is utilised on all datasets. 

\subsection{Metrics}

Fidelity measures the average change in the prediction by the removal of important nodes as identified by the explainability model. As a GCN is a transductive model, the graph structure cannot be changed. Important nodes are occluded by setting the nodal feature to zero. Fidelity is defined as follows:
\begin{definition}
    Let $\mathcal{G}$ be a graph, and denote the occluded graph by $\hat{\mathcal{G}}$. Fidelity is defined as
    \begin{equation}
        \mathrm{Fidelity} = \frac{1}{N}\sum_{i}^{N} F(\mathcal{G}_i)-F(\hat{\mathcal{G}}_i).
    \end{equation}
\end{definition}
The removal of important nodes results in a drop in the prediction probability corresponding to a high fidelity. Negative fidelity occurs when $f(\mathcal{G}) < f(\hat{\mathcal{G}})$. Negative fidelity implies that the explainability model has not identified important nodes as removing them corresponds to stronger prediction. 

An XAI model should highlight important features and provide explanations which are as simple as possible. Sparsity measures the number of nodes in an explanation relative to the size of the graph. Sparsity is defined as:
\begin{definition}
    Let $\mathcal{G}_i$ be a graph and $\mathcal{H}_i$ be the sub-graph of important nodes identified by an explainability model. 
    \begin{equation}
        \mathrm{Sparsity} = \frac{1}{N}\sum_{i}^{N} 1-\frac{|\mathcal{H}_i|}{|\mathcal{G}_i|}.
    \end{equation}
\end{definition}
Under Fidelity and Sparsity, an ideal explainability model requires a small number of nodes which result in a large drop in prediction probability when occluded.

Given a dataset with explanation group-truths,  graph explanation accuracy is measured by the Jaccard index. The Jaccard index measures the overlap between the ground-truth explanation graph and the important nodes identified by the explainability model. The Jaccard index is defined as:

\begin{definition}
    Let $M^{g}$ and $M^{p}$ be the ground-truth and predicted graph explanations, respectively. The Jaccard index is defined as: 
    \begin{equation}
    \mathrm{Jaccard}(M^{g}, M^{p}) = \frac{|M^{g} \cap M^{p}|}{|M^{g} \cup M^{p}|}.
\end{equation}
\end{definition}
Integrated gradients provides an attribution for all nodes. The sign of the attribution corresponds to which class the nodes or features move a classification. Defining important nodes for the metrics are not possible without a threshold for what an important attribution value is. We find in literature that a threshold is not specified for the metrics \cite{yuan2023explainability}. Further, direct comparison between explainability models are not possible as the attributions are on different scales. Explanations are normalised between 0 and 1 to allow for comparison on the same scale. Following \cite{Agarwal2023Evaluating} the threshold is set to 0.8. Further investigation of an appropriate threshold is left to future work.

\subsection{Synthetic and Real-World Datasets}

\begin{figure}[!b]
    \centering
    \begin{minipage}[b]{0.4\linewidth}
        \centering
        \includegraphics[width=\linewidth]{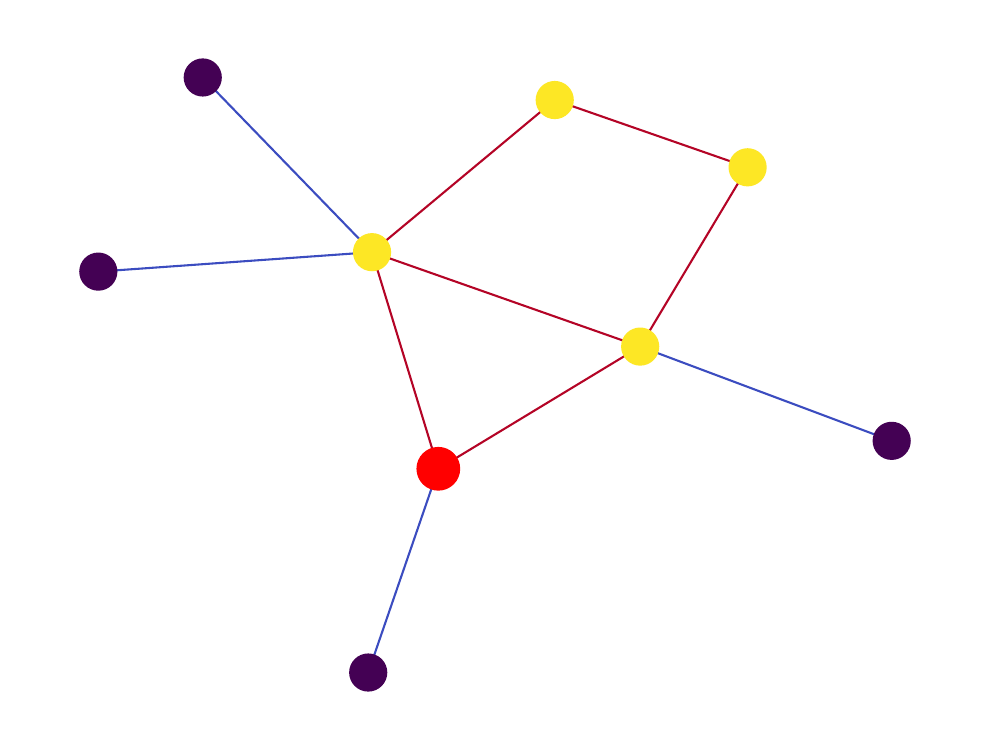}
        % \caption*{(a) `House'}
    \end{minipage}
    \hfill
    \begin{minipage}[b]{0.4\linewidth}
        \centering
        \includegraphics[width=\linewidth]{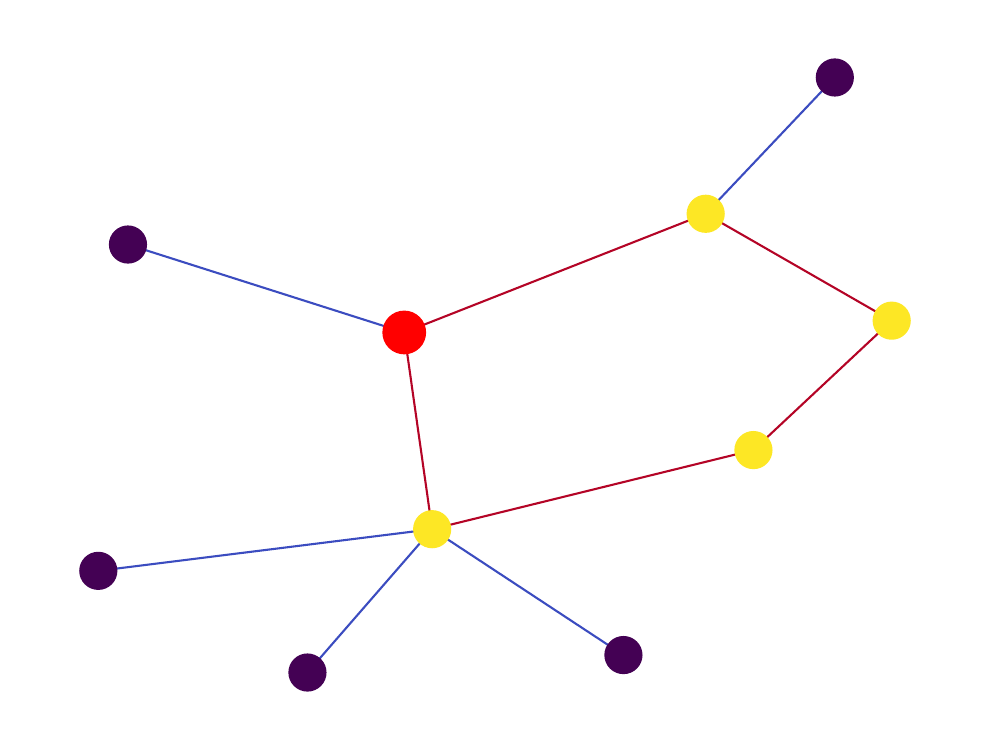}
        % \caption*{(b) `Circle'}
    \end{minipage}
    \caption{Example motifs from ShapeGGen library. Red: node to explain, Yellow: important nodes in the motif, Purple: non-important nodes in the graph. Edge colours are for edge importance which are not considered in this work. \textbf{Left}: House motif, \textbf{Right}: Circle motif.}
    \label{fig:motifs}
\end{figure}

ShapeGGen \cite{Agarwal2023Evaluating} is a graph explainability dataset generator. Graphs in ShapeGGen are built around a graph motif. In Figure \ref{fig:motifs} we provide two examples of motifs. Motifs in graphs have been identified as an important geometric aspect in classification tasks. Chemical compounds, for example, are classified by underlying motifs. ShapeGGen allows further changes to the geometry of a graph by altering: the homophily constant, number of clusters, number and size of subgraphs, and the probability that two nodes are connected.  

Four synthetic datasets are generated to assess the robustness of IG and GB-IG under changing geometry.
The homophily coefficient and motif of the graphs are altered to induce different geometries \cite{Agarwal2023Evaluating}.

The homophily coefficient determines if nodes with similar features and labels are connected. A homophily coefficient of -1 produces a heterophilic graph where nodes with dissimilar features and labels are connected \cite{Newman2010}. A homophily coefficient of 1 induces a graph in which nodes with similar features and labels connect. The homophily coefficient is altered as graph explainability models often suffer from a lack of robustness when homophily is altered \cite{Agarwal2023Evaluating}. 
Table \ref{table:synthetic} provides the results of GB-IG and IG with three common base-point choices evaluated on four synthetic datasets. GB-IG outperforms IG under fidelity and Jaccard. Table \ref{table:synthetic} indicates that GB-IG is able to identify the underlying motif and the features with the largest impact on node classification. IG-zero provides a negative fidelity across the datasets indicating that the features IG-zero identifies as important increase the confidence of the model when removed. Furthermore, IG-gaussian is unable to identify structurally important motifs across all datasets. The results in Table \ref{table:synthetic} demonstrate that GB-IG accurately identifies important structures and nodes under changing geometries compared other IG based techniques. 

On each synthetic dataset GB-IG has the lowest sparsity. Comparing GB-IG with the highest sparsity there is an average difference of 0.02 across the four datasets. Sparsity neglects the contents of the explanation. IG-gaussian has the highest average sparsity, but is unable to identify important motifs. Sparsity indicates the explanations provided by GB-IG are more complicated but the difference is negligible. 

We further evaluate our technique on three real-world datasets: Pubmed, Cora, CiteSeer. Table \ref{table:real} provides the performance of GB-IG and IG across the three datasets evaluated with fidelity and sparsity. The Jaccard index cannot be applied as the datasets do not have explanation ground-truths.

IG-zero and IG-uniform have a negative fidelity across the datasets in Table~\ref{table:real}. The negative fidelity indicates that the nodes identified as important by the explainability models are incorrect as the removal of the important nodes increases the prediction probability. IG-Gaussian marginally outperforms GB-IG on Cora with a difference in fidelity of 0.0012. On CiteSeer and Pubmed, GB-IG better identifies important nodes in a prediction than IG with varying base-points. 

Tables \ref{table:synthetic} and \ref{table:real} demonstrate IG-Gaussian has the best overall sparsity. The average difference between GB-IG and IG-Gaussian sparisty on the real world datasets in Table \ref{table:real} is 0.1. Whilst the explanations of IG-Gaussian may be simpler (highest sparsity), the explanations do not reveal important nodes in a classification, the true goal of explainable AI which GB-IG achieves. 

\begin{table}[!ht]
\centering
\caption{Evaluation of graph XAI models on four synthetic datasets generated by ShapeGGen library. Sparsity, Fidelity and Jaccard index are used as performance metrics. Bold indicates best performance.
}
\label{table:synthetic}
\begin{tabular}[]{l|c|c|c|c|c|c}
\toprule
Metric & Motif & Homophily & IG-Zero & IG-Uniform & IG-Gaussian & GB-IG (Ours) \\ \midrule

\multicolumn{1}{l|}{} & House & $ +1$ & $0.8422$ & $0.8463$ & $\mathbf{0.8568}$ & $0.8168$  \\
\multicolumn{1}{l|}{Sparsity} & House & $-1$ & $0.8368$ & $0.8398$ & $\mathbf{0.8371}$ & $0.8259$\\

\multicolumn{1}{l|}{} & Circle &$ +1$ & $0.8274$ & $\mathbf{0.8407}$ & $0.8375$ & $0.8111$\\
\multicolumn{1}{l|}{} & Circle & $-1$ & $0.8283$ & $0.8215$ & $\mathbf{0.8375}$ & $0.8168$\\
 \midrule
\multicolumn{1}{l|}{} & House & $+1$ & $-0.0554$ & $0.0212$ & $0.0137$ & $\mathbf{0.0233}$  \\
\multicolumn{1}{l|}{Fidelity} &  House & $-1$ & $-0.0400$ & $0.0267$ & $0.0339$ & $\mathbf{0.0557}$\\

\multicolumn{1}{l|}{} & Circle &$ +1$ & $-0.0375$ & $0.0189$ & $0.0190$ & $\mathbf{0.0422}$\\
\multicolumn{1}{l|}{} & Circle & $-1$ & $-0.1771$ & $-0.0611$ & $-0.0246$ & $\mathbf{0.0114}$\\
\midrule 
\multicolumn{1}{l|}{} & House & $+1$ & $0.0007$ & $0.0067$ & $0.0000$ & $\mathbf{0.1045}$ \\
\multicolumn{1}{l|}{Jaccard Index} & House & $-1$ & $0.0048$ & $0.0172$ & $0.0000$ & $\mathbf{0.1045}$\\
\multicolumn{1}{l|}{} & Circle &$+1$ & $0.0061$ & $0.0591$ & $0.0000$ & $\mathbf{0.2150}$\\
\multicolumn{1}{l|}{} & Circle & $-1$ & $0.0160$ & $0.0614$ & $0.0000$ & $\mathbf{0.2291}$\\

\bottomrule
\end{tabular}
\end{table}%

\begin{table}[!ht]
\centering
\caption{Evaluation of graph XAI models on three well-known node classification datasets. Sparsity and Fidelity used as performance metrics. The Jaccard index is not used as the datasets do not have ground truth. Bold indicates best performance.}
\label{table:real}
\begin{tabular}[]{l|c|c|c|c|c}
\toprule
Metric & Dataset & IG-Zero & IG-Uniform & IG-Gaussian & GB-IG (Ours) \\ \midrule

\multicolumn{1}{l|}{} & Pubmed& $-0.0515$ & $0.0000$ & $0.0027$ & $\mathbf{0.0177}$\\
\multicolumn{1}{l|}{Fidelity} & Cora & $-0.0557$ & $-0.0266$ & $\mathbf{0.0079}$ & $0.0067$\\
\multicolumn{1}{l|}{} & CiteSeer& $-0.0659$ & $-0.0180$ & $0.0042$ & $\mathbf{0.0132}$\\

 \midrule
 \multicolumn{1}{l|}{} & Pubmed& $ 0.9243$ & $0.0000$ & $\mathbf{0.9291}$ & $0.7828$\\

\multicolumn{1}{l|}{Sparsity} &  Cora  & $0.8837$ & $0.9059$ & $\mathbf{0.9104}$ & $0.8156$\\
\multicolumn{1}{l|}{} & CiteSeer& $0.8359$ & $0.8570$ & $ \mathbf{0.8718}$ & $0.7840$\\

\bottomrule
\end{tabular}
\end{table}%

\section{Conclusions and Future Works}
In this work, we introduced GB-IG, a novel extension of IG to graphs. We demonstrated that GB-IG outperforms IG in finding nodes and geometric structures important in node classification tasks. We validated our technique on four synthetic graph datasets with explanation ground-truths and three prevalent graph datasets. For future work, we will investigate limiting the number of paths in GB-IG. Reducing the number of paths will reduce computational complexity and increase sparsity.

%
%
%

\bibliographystyle{splncs04}
\bibliography{mybibliography}

\begin{thebibliography}{10}
\providecommand{\url}[1]{\texttt{#1}}
\providecommand{\urlprefix}{URL }
\providecommand{\doi}[1]{https://doi.org/#1}

\bibitem{Agarwal2023Evaluating}
Agarwal, C., Queen, O., Lakkaraju, H., Zitnik, M.: Evaluating {E}xplainability for {G}raph {N}eural {N}etworks. Scientific Data  \textbf{10} (2023)

\bibitem{brodt}
Bordt, S., Uddeshya, U., Akata, Z., von Luxburg, U.: {T}he {M}anifold {H}ypothesis for {G}radient-{B}ased {E}xplanations. In: 2023 IEEE/CVF Conference on Computer Vision and Pattern Recognition Workshops (CVPRW). pp. 3697--3702 (2023)

\bibitem{bronstein2021geometricdeeplearninggrids}
Bronstein, M.M., Bruna, J., Cohen, T., Veličković, P.: {G}eometric {D}eep {L}earning: {G}rids, {G}roups, {G}raphs, {G}eodesics, and {G}auges. ar{X}iv preprint ar{X}iv:2104.13478  (2021)

\bibitem{costanza2025}
Costanza, F., Simpson, L.: {R}iemannian {I}ntegrated {G}radients: {A} {G}eometric {V}iew of {E}xplainable {AI}. arXiv preprint arXiv:2503.00892  (2025)

\bibitem{AUS2025AIGuardrails}
{Department of Industry, Science and Resources}: Introducing {M}andatory {G}uardrails for {AI} in {H}igh-{R}isk {S}ettings (2025), \url{https://consult.industry.gov.au/ai-mandatory-guardrails}, accessed: 2025-05-28

\bibitem{drakard2022exploring}
Drakard, D., Liu, R., Yosinski, J.: {E}xploring {U}nfairness in {I}ntegrated {G}radients {B}ased {A}ttribution {M}ethods. {O}pen{R}eview  (2022)

\bibitem{EU2025AIAct}
{European Commission}: {AI} {A}ct (2025), \url{https://digital-strategy.ec.europa.eu/en/policies/regulatory-framework-ai}, accessed: 2025-05-28

\bibitem{fefferman2013testing}
Fefferman, C., Mitter, S., Narayanan, H.: {T}esting the {M}anifold {H}ypothesis. Journal of the American Mathematical Society  \textbf{29},  983–1049 (2016)

\bibitem{FOULADVAND2023104407}
Fouladvand, S., Gomez, F.R., Nilforoshan, H., Schwede, M., Noshad, M., Jee, O., You, J., Sosic, R., Leskovec, J., Chen, J.: {G}raph-based clinical recommender: {P}redicting specialists procedure orders using graph representation learning. Journal of Biomedical Informatics  \textbf{143},  104407 (2023)

\bibitem{Hallowell_Badger_Sauerbrei_Nellåker_Kerasidou_2022}
Hallowell, N., Badger, S., Sauerbrei, A., Nellåker, C., Kerasidou, A.: “{I} don’t think people are ready to trust these algorithms at face value”: {T}rust and the use of machine learning algorithms in the diagnosis of rare disease. BMC Medical Ethics  \textbf{23}(1) (2022)

\bibitem{Jha_Saha_Singh_2022}
Jha, K., Saha, S., Singh, H.: {P}rediction of protein–protein interaction using graph neural networks. Scientific Reports  \textbf{12}(1),  1–12 (2022)

\bibitem{kindermans2017unreliability}
Kindermans, P.J., Hooker, S., Adebayo, J., Alber, M., Schütt, K.T., Dähne, S., Erhan, D., Kim, B.: {T}he ({U}n)reliability of {S}aliency {M}ethods. arXiv preprint arXiv:1711.00867  (2017)

\bibitem{kipf2017semi}
Kipf, T.N., Welling, M.: {S}emi-{S}upervised {C}lassification with {G}raph {C}onvolutional {N}etworks. In: Proceedings of the 5th International Conference on Learning Representations (ICLR) (2017)

\bibitem{lundstrom2022rigorous}
Lundstrom, D., Huang, T., Razaviyayn, M.: {A} {R}igorous {S}tudy of {I}ntegrated {G}radients {M}ethod and {E}xtensions to {I}nternal {N}euron {A}ttributions. Proceedings of the 39th International Conference on Machine Learning  \textbf{162},  14485--14508 (2022)

\bibitem{9737249}
Millar, K., Simpson, L., Cheng, A., Chew, H.G., Lim, C.C.: {D}etecting {B}otnet {V}ictims {T}hrough {G}raph-{B}ased {M}achine {L}earning. In: 2021 International Conference on Machine Learning and Cybernetics (ICMLC). pp.~1--6 (2021)

\bibitem{Newman2010}
Newman, M.E.J.: {N}etworks: {A}n {I}ntroduction. Oxford University Press, Oxford; New York (2010)

\bibitem{Ni_Lin_Luo_Gao_2019}
Ni, C.C., Lin, Y.Y., Luo, F., Gao, J.: Community detection on networks with {R}icci flow. Scientific Reports  \textbf{9}(1) (2019)

\bibitem{Pope_2019_CVPR}
Pope, P.E., Kolouri, S., Rostami, M., Martin, C.E., Hoffmann, H.: {E}xplainability {M}ethods for {G}raph {C}onvolutional {N}eural {N}etworks. In: Proceedings of the IEEE/CVF Conference on Computer Vision and Pattern Recognition (CVPR) (2019)

\bibitem{NEURIPS2020_417fbbf2}
Sanchez-Lengeling, B., Wei, J., Lee, B., Reif, E., Wang, P., Qian, W., McCloskey, K., Colwell, L., Wiltschko, A.: {E}valuating {A}ttribution for {G}raph {N}eural {N}etworks. In: Advances in Neural Information Processing Systems. vol.~33, pp. 5898--5910 (2020)

\bibitem{sanyal-ren-2021-discretized}
Sanyal, S., Ren, X.: {D}iscretized {I}ntegrated {G}radients for {E}xplaining {L}anguage {M}odels. In: Proceedings of the 2021 Conference on Empirical Methods in Natural Language Processing. pp. 10285--10299. Association for Computational Linguistics (2021)

\bibitem{10935253}
Simpson, L., Costanza, F., Millar, K., Cheng, A., Lim, C.C., Chew, H.G.: {A}lgebraic {A}dversarial {A}ttacks on {I}ntegrated {G}radients. 2024 International Conference on Machine Learning and Cybernetics (ICMLC) pp. 26--31 (2024)

\bibitem{simpson-basepoint}
Simpson, L., Costanza, F., Millar, K., Cheng, A., Lim, C.C., Chew, H.G.: {T}angentially {A}ligned {I}ntegrated {G}radients for {U}ser-{F}riendly {E}xplanations. Irish Conference on Artificial Intelligence and Cognitive Science (AICS) pp. 1--11 (2025)

\bibitem{sturmfels2020visualizing}
Sturmfels, P., Lundberg, S., Lee, S.I.: {V}isualizing the {I}mpact of {F}eature {A}ttribution {B}aselines. Distill  (2020), https://distill.pub/2020/attribution-baselines

\bibitem{pmlr-v70-sundararajan17a}
Sundararajan, M., Taly, A., Yan, Q.: {A}xiomatic {A}ttribution for {D}eep {N}etworks. Proceedings of the 34th International Conference on Machine Learning (ICML)  \textbf{70},  3319--3328 (2017)

\bibitem{taha2015metrics}
Taha, A.A., Hanbury, A.: {M}etrics for {E}valuating {3D} {M}edical {I}mage {S}egmentation: {A}nalysis, {S}election, and {T}ool. BMC Medical Imaging  \textbf{15}, ~29 (2015)

\bibitem{whiteley2024statisticalexplorationmanifoldhypothesis}
Whiteley, N., Gray, A., Rubin-Delanchy, P.: {S}tatistical {E}xploration of the {M}anifold {H}ypothesis. arXiv preprint arXiv:2208.11665  (2024)

\bibitem{pmlr-v196-xenopoulos22a}
Xenopoulos, P., Chan, G., Doraiswamy, H., Nonato, L.G., Barr, B., Silva, C.: {GALE}: {G}lobally {A}ssessing {L}ocal {E}xplanations. In: Proceedings of Topological, Algebraic, and Geometric Learning Workshops 2022. Proceedings of Machine Learning Research (PMLR), vol.~196, pp. 322--331 (2022)

\bibitem{yuan2023explainability}
Yuan, H., Yu, H., Gui, S., Ji, S.: {E}xplainability in {G}raph {N}eural {N}etworks: {A} {T}axonomic {S}urvey. IEEE Transactions on Pattern Analysis and Machine Intelligence  \textbf{45}(5),  5782--5799 (2023)

\bibitem{zaher2024manifoldintegratedgradientsriemannian}
Zaher, E., Trzaskowski, M., Nguyen, Q., Roosta, F.: {M}anifold {I}ntegrated {G}radients: {R}iemannian {G}eometry for {F}eature {A}ttribution. arXiv preprint arXiv:2405.09800  (2024)

\bibitem{zednik2019solving}
Zednik, C.: {S}olving the {B}lack {B}ox {P}roblem: {A} {N}ormative {F}ramework for {E}xplainable {A}rtificial {I}ntelligence. Philosophy \& Technology  \textbf{34}(2),  265--288 (2021)

\end{thebibliography}

\end{document}